\documentclass{article}

\usepackage[final,nonatbib]{neurips_2023}

\usepackage[utf8]{inputenc} % allow utf-8 input
\usepackage[T1]{fontenc}    % use 8-bit T1 fonts
\usepackage{hyperref}       % hyperlinks
\usepackage{url}            % simple URL typesetting
\usepackage{booktabs}       % professional-quality tables
\usepackage{amsfonts}       % blackboard math symbols
\usepackage{nicefrac}       % compact symbols for 1/2, etc.
\usepackage{microtype}      % microtypography
\usepackage{xcolor}         % colors
\usepackage{biblatex}
\usepackage{graphicx}
\usepackage{amsmath, bm}
\newcommand{\dd}{\text{d}}

\renewcommand{\l}{\left}
\renewcommand{\r}{\right}

\title{Using Ornstein--Uhlenbeck Process to understand Denoising Diffusion Probabilistic Model and its Noise Schedules}

\author{Javier E.~Santos\\
Computational Earth Science Group (EES-16), Earth and Environmental Sciences Division\\
Los Alamos National Laboratory, Los Alamos, NM 87545, USA
\And
Yen Ting Lin\\
Information Sciences Group (CCS-3), Computational and Statistical Sciences Division\\
Los Alamos National Laboratory, Los Alamos, NM 87545, USA\\
\texttt{yentingl@lanl.gov}}

\begin{document}

\maketitle

\begin{abstract}
The aim of this short note is to show that Denoising Diffusion Probabilistic Model DDPM, a non-homogeneous discrete-time Markov process, can be represented by a time-homogeneous continuous-time Markov process observed at non-uniformly sampled discrete times. Surprisingly, this continuous-time Markov process is the well-known and well-studied Ornstein--Ohlenbeck (OU) process, which was developed in 1930's for studying Brownian particles in Harmonic potentials. We establish the formal equivalence between DDPM and the OU process using its analytical solution. We further demonstrate that the design problem of the noise scheduler for non-homogeneous DDPM is equivalent to designing observation times for the OU process. We present several heuristic designs for observation times based on principled quantities such as auto-variance and Fisher Information and connect them to \emph{ad hoc} noise schedules for DDPM. Interestingly, we show that the Fisher-Information-motivated schedule corresponds exactly the \emph{cosine schedule}, which was developed without any theoretical foundation but is the current state-of-the-art noise schedule. 
\end{abstract}

\section{Denoising Diffusion Probabilistic Model}

We begin with a concise description of the Denoising Diffusion Probabilistic Models (DDPM) \cite{sohl-dicksteinDeepUnsupervisedLearning2015,hoDenoisingDiffusionProbabilistic2020}. DDPM is a non-homogeneous discrete-time Markov chain defined by 
\begin{equation}
     \mathbf{Z}_{k} = \sqrt{\alpha_k } \mathbf{Z}_{k-1} + \sqrt{1 - \alpha_k } \boldsymbol{\varepsilon}_{k-1}, \label{eq:DDPM}
\end{equation}
where $k=0,1\ldots T$ is the discrete-time index, $\mathbf{Z}_k$ is a rank-1 vector, $\boldsymbol{\varepsilon}_k \sim \mathcal{N}\l(0,\mathbf{I}\r)$ is a standard multivariate Gaussian random variable, and $\alpha_k \in [0,1)$. %The initial condition, $\mathbf{x}_0$ is a vector of flattened intensity information of an image. 
As shown in \cite{sohl-dicksteinDeepUnsupervisedLearning2015} and \cite{hoDenoisingDiffusionProbabilistic2020}, the solution to the above discrete-time Markov process is
\begin{equation}
    \mathbf{Z}_{k} = \sqrt{\prod_{i=1}^k \alpha_i } \mathbf{Z}_{0} + \sqrt{1 - \prod_{i=1}^k \alpha_i } \bar{\boldsymbol{\varepsilon}_k}, \label{eq:DDPMsol}
\end{equation}
where $\bar{\boldsymbol{\varepsilon}_k}\sim \mathcal{N}\l(0,\mathbf{I}\r)$ is another standard multivariate Gaussian random variable. The term $\prod_{i=1}^k \alpha_i$ is often denoted by $\bar{\alpha}_k$ in the literature. 

DDPM is a non-homogeneous process because the strength of the noise, quantified as $1-\alpha_k$, is not the same for each discrete time $k$. It is known that the performance of the DDPM depends critically on the setup of $\l\{\beta_k \r\}_k:=\l\{1-\alpha_k \r\}_k$; see for example \cite{nicholImprovedDenoisingDiffusion2021}. Although $\mathbf{X}$ is a vector, it is sufficient for us to consider one of the components below, because each channel is transformed by independent Gaussian random numbers ($\boldsymbol{\epsilon}_{k}$ has a diagonal covariance matrix).

\section{Ornstein--Uhlenbeck Process and its Analytical Solution}
%Let us switch gear and consider the Ornstein--Uhlenbeck (OU) process 
Let's shift our focus to the Ornstein--Uhlenbeck (OU) process \cite{originalOU,vanKampen,gardinerStochasticMethodsHandbook2009}, which can described by the It\^o stochastic differential equation (SDE):
\begin{equation}
    \dd X_t = - \gamma X_t \dd t + \sigma \dd W_t, 
\end{equation}
where $\gamma$ is the relaxation rate, $\sigma$ is the strength of fluctuation, and $W_t$ is the standard Wiener process. OU is a time-homogeneous process because $\gamma$ and $\sigma$ are time-invariant. 

%It is elementary to solve the It\^o SDE by the following steps. 
It is straightforward to solve the Itô SDE using the following steps. We first transform the stochastic process into the moving frame of the eigensystem: $Y_t := X_t e^{\gamma t}$, whose evolutionary equation can be derived by applying the It\^o formula to $Y_t$. The resulting evolutionary equation is $\dd Y_t =  \sigma e^{\gamma t} \dd W_t$. After the integration from time $s$ to $t>s$, we arrived at the solution $Y_t = Y_s + \sigma \sqrt{\l(e^{2 \gamma (t-s)} - 1\r)/\l(2\gamma\r) } W_1$, which can be transformed back to the solution of $X_t$:
\begin{equation}
    X_{t} = X_s e^{-\gamma (t-s)} + \sigma  \sqrt{\frac{1}{2 \gamma} \l(1 - e^{-2 \gamma (t-s)}\r) } W_1, \quad t,\, s \ge 0. \label{eq:OUsol}
\end{equation}
Note that $W_1$ is the standard Wiener process at $t=1$, which is a standard Gaussian random variable. 

\section{Equivalence between Denoising Diffusion Probabilistic Model and Ornstein--Uhlenbeck Process}

Setting $\gamma=1$ and $\sigma=\sqrt{2}$ and comparing Eqs.~\eqref{eq:OUsol} to \eqref{eq:DDPMsol}, we can establish that DDPM is equivalent to OU process observed at discrete times in this parametrization. That is, the joint probability distribution of the OU process at these specified times is identical to that of the DDMP. Suppose we define a set of ordered observation times $\l\{t_k\r\}_k$:
\begin{align}
    t_k ={}& - \frac{1}{2} \sum_{i=1}^k \log \alpha_i. \label{eq:major}
\end{align}
Note that the discrete times $\l\{t_k\r\}_k$ are uniquely determined by the noise schedule $\l\{\beta_k\r\}_k $, and vice versa.
We claim $Z_k=X_{t_k}$. To see this, we set $t=t_{k}$ and $s=t_{k-1}$ specified by Eq.~\eqref{eq:major}, and then Eq.~\eqref{eq:OUsol} reduces to the Markov evolutionary equation \eqref{eq:DDPM} of  DDPM: 
\begin{equation}
    Z_{k} = X_{t_{k}} = X_{t_{k-1}} e^{- \frac{1}{2}\log \alpha_k } + \sqrt{1-e^{-\log \alpha_k }} W_1 \equiv  \sqrt{\alpha_k} Z_{k-1} + \sqrt{1-\alpha_k} \varepsilon_{k-1}. 
\end{equation}
Similarly, by setting $t=t_{k}$ and $s=t_0=0$, the OU solution Eq.~\eqref{eq:OUsol} reduces to the DDPM solution (Eq.~\eqref{eq:DDPMsol}):
\begin{align}
    Z_{k} = X_{t_{k}} ={}& X_{t_0} e^{- \frac{1}{2} \sum_{i=1}^k \log \alpha_i } + \sqrt{1-e^{-\log  \sum_{i=1}^k \log \alpha_i  }} W_1 \nonumber \\ \equiv {}&  \sqrt{\prod_{i=1}^k \alpha_i} Z_{0} + \sqrt{1-\prod_{i=1}^k \alpha_i} \varepsilon_{k-1}
\end{align}

As we are using the standard It\^o SDE form to describe the OU process, one wonders if this identification has been already made by existing continous-time It\^o SDE form of Generative Diffusion Models \cite{songScoreBasedGenerativeModeling2021a}. We here remark the difference. Indeed, \cite{songScoreBasedGenerativeModeling2021a}~also claimed their proposed SDE includes the DDPM \cite{songScoreBasedGenerativeModeling2021a}. However, their argument was based on an It\^o process whose noise strength $\sigma(t)$ is \emph{time-dependent}; the claimed equivalence would require infinitely many steps in the DDMP (i.e., $k=0\ldots N$, $N\rightarrow \infty$). In a recent paper by \cite{DDS}, the authors also adopted the term \emph{Ornstein--Uhlenbeck process}, but similar to \cite{songScoreBasedGenerativeModeling2021a}, the process is not the standard time-homogeneous OU process \cite{originalOU}: the process in \cite{DDS} has time-dependent drift and diffusion. In contrast, our identification of the equivalence here does not depend on these conditions: for any $\l\{\alpha_k\r\}_{k=1}^T$ defined in a DDPM, Eq.~\eqref{eq:major} identifies the corresponding discrete times of the standard, time-homogeneous, OU process. \\

\section{Noise schedule as a design problem of observational times}

The analytical solution of the OU process provides a way to probe the design principle of different noise schedules in terms of the same process. First, because we have the analytical solution of the auto-variance of the process:
\begin{equation}
    \text{var}\left[X_t \right] = 1-e^{-2 t},
\end{equation}
let us first consider a set of $T$ observation times that between two observations the change of auto-variance is constant:
\begin{equation}
    e^{-2 t_{k-1}}-e^{-2 t_k} = \mathcal{C},\, k = 1,2,\ldots T,
\end{equation}
where $\mathcal{C}$ is a constant. Noting that by $e^{-2 t_{k}}=\bar{\alpha}_k \equiv \prod_{i=1}^k \alpha_i$, the above condition leads to
\begin{equation}
    \bar{\alpha}_{k-1} - \bar{\alpha}_k = \mathcal{C}, 
\end{equation}
and assuming that at the discrete time $T$ the system evolves to OU's limiting distribution $t\rightarrow \infty$, 
\begin{equation}
    \bar{\alpha}_k = 1- \frac{k}{T}.
\end{equation}
This leads to
\begin{equation}
    \alpha_k = \frac{T-k}{T-k+1} \text{ and } \beta_k = \frac{1}{T-k+1}, 
\end{equation}
which is the noise schedule in \cite{sohl-dicksteinDeepUnsupervisedLearning2015}. As such, \emph{we can understand the noise schedule is chosen so that the auto-variances are steadily increasing (as DDPM's discrete time increases)}. 

As the second heuristic method, let us consider the coefficient of variation (CV):
\begin{equation}
    \text{CV}\equiv \frac{\sqrt{\text{var}\left[X_t\right]}}{\mathbb{E}\left[X_t\right] }=\frac{\sqrt{1-e^{-2t}}}{e^{-t}}.
\end{equation}
CV is related to the signal-to-noise ratio (SNR), which is often defined as $\text{SNR}\equiv {\mathbb{E}\left[X_t\right] }^2/\text{var}\left[X_t\right]=\text{CV}^{-2}$. The idea is that we would like to put more resources in those time regions where the CV is low (or equivalently, where the SNR is high). Introducing a parameter $\theta=e^{-t} \in (0,1]$, we can formulate a desired sampling probability density function
\begin{equation}
    \pi(\theta) =  \text{CV}^{-1}(\theta) = \frac{\theta}{\sqrt{1-\theta^2}},
\end{equation}
whose cumulative distribution function can be derived:
\begin{equation}
    \text{CDF}(\theta) = \sqrt{1-\theta^2}.
\end{equation}
Here, we follow the convention of the forward process moving from $t:0\rightarrow \infty$, thus $\theta:1\rightarrow 0$. 
The CDF can be used for inverse transform sampling: we consider $T+1$ evenly distributed CDF marks between $(0,1)$, that $c_k = k/T$, $k=0, \ldots T$. The corresponding $\theta_k$ is
\begin{equation}
    c_k = \frac{k}{T} = 1 - \sqrt{1-\theta_k^2}
\end{equation}
or equivalently,
\begin{equation}
    1 - \theta_k^2 = 1- \frac{k^2}{T^2}.
\end{equation}
Because $\theta_k^2 = e^{-2 t_k} = \bar{\alpha}_k$, \emph{this heuristic is equivalent to a quadratic program of $\bar{\alpha}$}, or equivalently, 
\begin{equation}
    \alpha_k = \frac{ \bar{\alpha}_k}{  \bar{\alpha}_{k-1}} = \frac{T^2-k^2}{T^2-(k-1)^2} \text{ and } \beta_k = 1- \alpha_k = \frac{2k-1}{T^2-(k-1)^2}.
\end{equation}

The third example involves using the differential entropy of the OU process:
\begin{equation}
    S(t) = \int_{-\infty}^\infty \rho\left(X_t = x\right) \log \rho\left(X_t = x \right) = \frac{1}{2} + \frac{1}{2} \log 2 \pi \left(1-e^{-2 t}\right)
\end{equation}
Note the apparent singularity at $t=0$, due to the nature of an initial $\delta$-distribution. We can, however, consider that the initial state has an intrinsic uncertainty due to the discretization of the pixel intensities. Suppose we uniformly distribute the initial intensity in between the discretized values and standardize the dataset, this initial uncertainty has a variance $\sigma_0^2$ of order $\mathcal{O}((1/257)^2 (1/12))$. We can model this uncertainty by only considering the OU process after $ t_0 = -1/2 \log (1-\sigma_0^2)$, hence bypassing the singularity at $t=0$. Next, we notice that $S\left(t\rightarrow \infty\right)=\left(1+\log 2 \pi\right)/2$, showing that the whole process creates $-\log (\sigma_0^2)$ entropy. This heuristic method aims to identify observation times with identical entropic production between consecutive observation times. As such, the discrete times can be solved by 
\begin{equation}
    S(t_k) - S(t_0) = \frac{1}{2} \log \frac{1-e^{-2t_k}}{\sigma_0^2} = \frac{k}{2T} \log \frac{1}{\sigma_0^2},
\end{equation}
which leads to 
\begin{equation}
    \bar{\alpha}_k=e^{-2t_k} = 1- \sigma_0^{2 -\frac{k}{T}},
\end{equation}
or equivalently
\begin{equation}
    {\alpha}_k=\frac{1- \sigma_0^{2 -\frac{k}{T}}}{1- \sigma_0^{2 -\frac{k-1}{T}}} \text{ and } {\beta}_k=\frac{\sigma_0^{-\frac{k}{T}}- \sigma_0^{-\frac{k-1}{T}}}{\sigma_0^{-2}- \sigma_0^{ -\frac{k-1}{T}}}.
\end{equation}

As the final example, let us consider a heuristic design of the observation times by Fisher Information of the process. Recall that in the implementation of DDPM, we are largely estimating the conditional mean of the reverse-time process. This is achieved by minimizing the $L^2$-loss of the prediction and the analytical solution, weighted by $1/\sigma^2 \approx \text{var}\left[X_t\right]$. Recall that the Fisher Information of the mean of the process Eq.~\eqref{eq:OUsol} at time $t$ is also $\text{var}\left[X_t\right]$. This observation motivates us to sample with a density $\pi$, proportional the square root of the Fisher Information:
\begin{equation}
    \pi(\theta) \propto \sqrt{\text{FI}(X_t)} = \frac{1}{\sqrt{1-\theta^2}},  \label{eq:FIdensity}
\end{equation}
where $\theta\equiv e^{-t}$ as defined before. We remind the reader of the definition of Fisher Information:
\begin{equation}
    \text{FI}(X_t) = \int_{-\infty}^\infty \rho\left(X_t = x; \theta \right) \left(\frac{\partial \rho\left(X_t = x;\theta\right)}{\partial \theta}\right)^2.
\end{equation}
Here, for some parameter $\theta$ of the process. Using the square root of the Fisher Information is natural because it is the only choice that the defined density is invariant under a change of variable. For example, consider a change of variable from $\theta=e^{-t}$ to the physical time $t=-\log(\theta)$:
\begin{align}
    \pi(\theta) \dd \theta ={}& \sqrt{\int_{-\infty}^\infty \rho\left(X_t = x; e^{-t} \right) \left(\frac{\partial \rho\left(X_t = x;e^{-t}\right)}{\partial t}\right)^2 \left(- \frac{\partial \log \theta}{\partial \theta}\right)^2}  \frac{\dd e^{-t}}{\dd t} \dd t \nonumber \\
    ={}& \sqrt{\int_{-\infty}^\infty \rho\left(X_t = x; t \right) \left(\frac{\partial \rho\left(X_t = x;t\right)}{\partial t}\right)^2} \dd t = \pi \left( t \right) \dd t.
\end{align}
From Eq.~\eqref{eq:FIdensity}, we can determine the normalization constant
\begin{equation}
    \mathcal{Z} = \int_{0}^1  \frac{1}{\sqrt{1-\theta^2}}  \dd \theta= \arcsin(1)-\arcsin(0) = \frac{\pi}{2}, 
\end{equation}
and the cumulative distribution function 
\begin{equation}
    \text{CDF}(\theta) = \int_\theta^1  \frac{1}{\sqrt{1-\theta^2}} \dd \theta = \frac{1}{\mathcal{Z}} \left[\arcsin(1) - \arcsin(\theta)\right] = 1 - \frac{2}{\pi} \arcsin(\theta),
\end{equation}
which is used for inverse transform sampling: we consider $T+1$ evenly distributed CDF marks between $(0,1)$, that $c_k = k/T$, $k=0, \ldots T$. The corresponding $\theta_k$ is
\begin{equation}
    \frac{k}{T} = 1 - \frac{2}{\pi} \arcsin(\theta_k),
\end{equation}
or equivalently 
\begin{equation}
     \theta_k = \sin\left(\frac{\pi}{2}\left(1-\frac{k}{T}\right)\right) = \cos\left(\frac{k\pi}{2T}\right).
\end{equation}
Recall that $\theta_k=e^{-t_k}$ and $\bar{\alpha}_k = e^{-2t_k}$ (i.e., Eq.~\eqref{eq:major})
\begin{equation}
     \bar{\alpha}_{k} = \cos^2\left(\frac{k\pi}{2T}\right),
\end{equation}
which is the cosine schedule \cite{nicholImprovedDenoisingDiffusion2021}. As such, \emph{we concluded that in contrast to the common belief that the cosine schedule is an \emph{ad hoc} engineered construct, it may have come from a deeper information-theoretic principle}.

\section{Numerical Experiments}
\begin{figure}[h!]
\centering
   \includegraphics[width=0.99\textwidth]{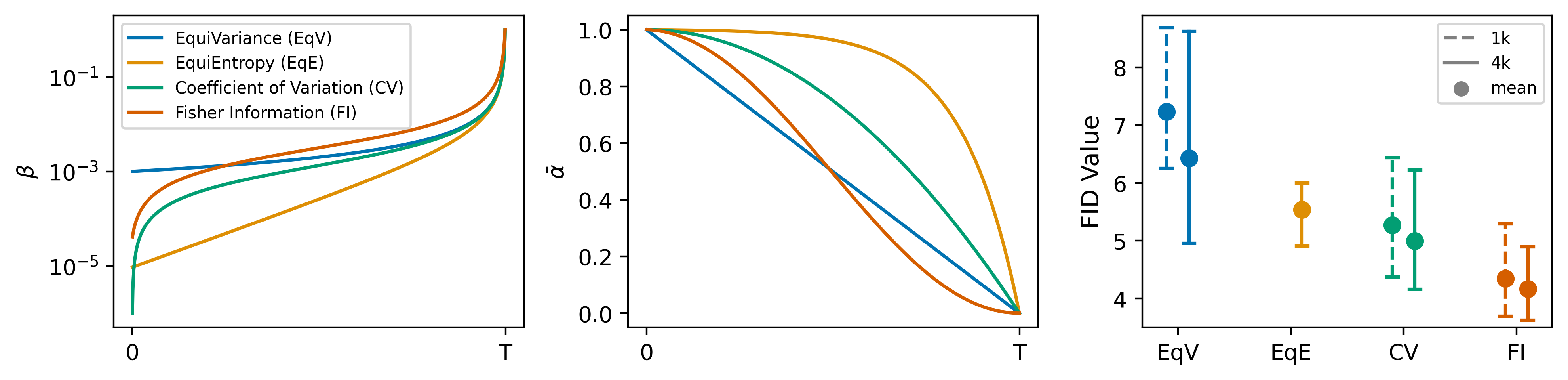}
    \caption{$\beta$ schedule (Left panel) and the corresponding $\bar{alpha}$ (Middle panel). (Right Panel) FID for 50,000 generated images. Error bars represent the maximum, average, and minimum across 5 checkpoints (saved every 100,000 iterations).  }
    \label{fig:FIDs}
\end{figure}

To demonstrate the effectiveness of various schedules, we conducted tests on CIFAR-10. We employed the model proposed in \cite{nicholImprovedDenoisingDiffusion2021} with both 1k and 4k steps using each noise schedule. We trained each model for 500,000 iterations, saving a checkpoint every 100k. From these checkpoints, we sampled 50k images and computed their FID (Fréchet Inception Distance). The results are illustrated in Figure \ref{fig:FIDs}.

\section{Discussion}

We illustrated that the time-inhomogeneous and discrete-time DDPM is mathematically equivalent to the time-homogeneous and continuous-time Ornstein--Uhlenbeck process observed at non-uniformly distributed observation times. We should remark that in the context of continuous-time and discrete-state models, \cite{campbell} also identified that a transformation of time can transform a time-inhomogeneous process to a homogeneous one. However, it is beyond the authors' knowledge that the exact equivalence between DDPM to the well-known Ornstein--Uhlenbeck process has been established.

During the review of this note, submitted as a contribution to NeurRIPS 23 Workshop on Diffusion Models, we received negative feedback from an anonymous reviewer, stating that they ``imagine'' the authors in \cite{songScoreBasedGenerativeModeling2021a} already knew our claim and did not feel the need to transform the time-inhomogeneous process to a homogeneous one, the equivalence between DDPM and time-homogeneous Ornstein--Uhlenbeck process is ``well-known'' in the literature (without providing references), and the rest of the paper is ``rather minor''. We disagree with the reviewer's assessment and humbly think it is still useful to explicitly state and document this in the public domain. We will leave this for the reader to decide if this is in any sense novel or useful.

We showcased a few heuristic designs based on the property of the Ornstein--Uhlenbeck process. In the derivation, we found that it is most natural to directly model $\bar{\alpha}_k$, which has a direct connection to the noise strength of the OU solution \eqref{eq:OUsol}. This is in contrast to most common approaches which model $\beta_k$ (e.g., linear or quadratic schedules in \cite{hoDenoisingDiffusionProbabilistic2020}, exponential schedules in \cite{campbell}, geometric progression in \cite{NEURIPS2020_92c3b916}). We argue that it is not the most natural way to model $\beta_k$, albeit a convenient one in engineering. Let us consider the simplest linear schedule \cite{hoDenoisingDiffusionProbabilistic2020} $\beta_k:=\beta_0 + (\beta_1 - \beta_0) k / T$. The resulting forward process critically depends on the choice of $T$, without any scaling relationship nor continuum limit $T\rightarrow \infty$. That is, consider $T'= M T$ with some scaling parameter $M\in \mathbb{N}$, there exists no scaling relationship between $\bar{\alpha}_k$ and $\bar{\alpha}_k'$ for the corresponding time step $k$ and $k'= M k$ between the two process, although $\beta_k = \beta_k'$: 
\begin{align}
    \bar{\alpha}_{k} ={}& \prod_{i=1}^{k} (1-\beta_i), \quad \bar{\alpha}_{k'} =\prod_{i=1}^{k} (1-\beta_{i'}).
\end{align}
In contrast, with any of the derivations we presented above, direct modeling of $\bar{\alpha}_k$ leads to nice scaling property $\bar{\alpha}_{k'} \approx \bar{\alpha}_{Mk}$.

We also tried to analyze an information-theoretic motivated design principle proposed by \cite{austin} on continuous-time and discrete-state models. Unfortunately, the method based on mutual information cannot be applied to continuous-state OU, as the key quantities would diverge due to the $\delta$-distribution. We document the analysis in Appendix \ref{app:mi} 

Perhaps the most interesting discovery of this research note is the identification of the Fisher-Information-motivated observation times leads to the cosine schedule, which was deemed as an \emph{ad hoc} design even by the original designer \cite{nicholImprovedDenoisingDiffusion2021}. 

We finally remark that all these one-dimensional designs based on the process on a single channel can at best be described as heuristic approaches. As the neural network is given multiple channels to learn the reverse-time (denoising) process, the information, encoded as correlation between channels, needs to be considered for a rational design of the noise schedules for Generative Diffusion Models. Nevertheless, such correlation is of course data-dependent and difficult to analyze. At the very least, it is our hope that the identification of the formal equivalence between DDPM and OU may pave the way for this challenging task. %At the very least, we could conclude with a hypothesis that it is not rational to expect the existence of an optimal one-for-all noise schedule. 

\acksection
The authors were supported by Laboratory Directed Research \& Development projects ``Diffusion Modeling with Physical Constraints for Scientific Data (20240074ER)'' and ``Uncertainty Quantification for Robust Machine Learning'' (20210043DR). 

\appendix

\section{Heuristic Design based on Mutual Information}\label{app:mi}
Following the proposition by \cite{austin}, we compute the mutual information of OU process, i.e., 
\begin{equation}
    I_{x_t;x_0}=S_{x_t}+S_{x_0}-S_{x_t,x_0}. 
\end{equation}

As usual, we assume 
\begin{equation}
    x_0 \sim \mathcal{N}\left(\alpha, \sigma_0^2\right) 
\end{equation}
where $\sigma_0 \sim \mathcal{O}((2/256)^2 (1/12))$ is the initial dispersion.  From the solution of OU, we know
\begin{equation}
    x_t\vert x_0 \sim \mathcal{N}\left(x_0, 1-\theta^2 \right) := \mathcal{N}\left(x_0, 1-\theta^2\right). 
\end{equation}

The PDF of joint distribution can be expressed as
\begin{equation}
    p(x_t, x_0) = \frac{1}{2\pi \sigma_0 \sqrt{\left(1-\theta^2\right)} } e^{-\frac{\left(x_0-\alpha\right)^2}{2\sigma_0^2}-\frac{\left(x_t-x_0\theta\right)^2}{2\left(1-\theta^2\right)}},
\end{equation}
which, has an entropy
\begin{align}
    S_{x_t, x_0} ={}& \int_{-\infty}^\infty \dd x_0 \int_{-\infty}^\infty \dd x_t \left[\log \left(2\pi\sigma_0\sqrt{\left(1-\theta^2\right)}\right) +\frac{\left(x_0-\alpha\right)^2}{2\sigma_0^2}+\frac{\left(x_t-x_0\theta\right)^2}{2\left(1-\theta^2\right)} \right] p(x_t,x_0) \nonumber \\
    ={}& \frac{1}{2} \log \left(2^2\pi^2\sigma_0^2 \left(1-\theta^2\right)\right) + 1 
\end{align}
As for the initial distribution, which does not depend on the final, has the entropy
\begin{equation}
    S_{x_0} = \frac{1}{2} \log \left(2 \pi \sigma_0^2  \right) + \frac{1}{2}.
\end{equation}
The only non-trivial one is the entropy of $x_t$, which requires marginalization over $x_0$:
\begin{align}
    p(x_t) ={}& \frac{1}{2 \pi \sigma_0\sqrt{\left(1-\theta^2\right)}} \int_{-\infty}^{\infty}  \exp\left\{-\frac{\left(x_0-\alpha\right)^2}{2\sigma_0^2}-\frac{\left(x_t-x_0\theta\right)^2}{2\left(1-\theta^2\right)}\right\} \dd x_0 \nonumber \\
    % ={}& \frac{1}{2 \pi \sigma_0\sqrt{\left(1-\theta^2\right)}} \int_{-\infty}^{\infty}  \exp\left\{-\frac{\left(x_0^2-2x_0 \alpha + \alpha^2\right)\left(1-\theta^2\right) + \sigma_0^2 \left(x_t^2 -2 x_t x_0 \theta + x_0^2 \theta^2\right)}{2\sigma_0^2\left(1-\theta^2\right)}\right\} \dd x_0  \\
    % ={}& \frac{1}{2 \pi \sigma_0\sqrt{\left(1-\theta^2\right)}} \int_{-\infty}^{\infty}  \exp\left\{-\frac{\left(\left(1-\theta^2 +\sigma_0^2 \theta^2\right)x_0^2-2x_0 \left[ \alpha \left(1-\theta^2\right) + x_t \theta \sigma_0^2 \right]+ \alpha^2\left(1-\theta^2\right) + \sigma_0^2 x_t^2 \right) }{2\sigma_0^2\left(1-\theta^2\right)}\right\} \dd x_0 \\
    ={}& \frac{1}{\sqrt{2 \pi \left(1-\theta^2 + \sigma_0^2 \theta^2\right)}} \exp \left\{- \frac{\alpha^2\left(1-\theta^2\right) + \sigma_0^2 x_t^2  - \frac{\left[ \alpha \left(1-\theta^2\right) + x_t \theta \sigma_0^2\right]^2}{1-\theta^2 +\sigma_0^2 \theta^2 }  }{2\sigma_0^2\left(1-\theta^2\right)}\right\} \nonumber \\
    ={}& \frac{1}{\sqrt{2 \pi \left(1-\theta^2 + \sigma_0^2 \theta^2\right)}} \exp \left\{-\frac{x_t^2}{1-\theta^2 + \sigma_0^2 \theta^2 }+ \ldots \right\}, 
\end{align}
whose entropy is 
\begin{equation}
    S_{x_t} = \frac{1}{2} \log \left(2 \pi \left(1-e^{-2t} + \sigma_0^2 e^{-2t}\right) \right) + \frac{1}{2}.
\end{equation}
So, 
\begin{align}
    I_{x_t;x_0}={}& S_{x_t}+S_{x_0}-S_{x_t,x_0} \nonumber \\
    %={}& \frac{1}{2} \log \left(2 \pi \left(1-e^{-2t} + \sigma_0^2 e^{-2t}\right) \right) + \frac{1}{2} + \frac{1}{2} \log \left(2 \pi \sigma_0^2  \right) + \frac{1}{2} - \frac{1}{2} \log \left(2^2\pi^2\sigma_0^2 \left(1-e^{-2t}\right)\right) - 1 \nonumber \\
    = {}& \frac{1}{2} \log \frac{\left(1-e^{-2t} + \sigma_0^2 e^{-2t}\right)  \sigma_0^2 }{\sigma_0^2 (1-e^{-2t})} = \frac{1}{2} \log \left( 1 + \sigma_0^2\frac{e^{-2t}}{1-e^{-2t}}\right).
\end{align}
Note that there's a singularity at $t=0$ due to the singular ($\delta(x_t-x_0)$) distribution. 

The discretization based on the mutual information, based on \cite{austin}, is to solve:
\begin{equation}
    \frac{k}{T} = 1 - \frac{I_{x_t;x_0}}{S_{x_0}}= 1 - \frac{\log \left( 1 + \sigma_0^2\frac{e^{-2t_i}}{1-e^{-2t_i}}\right)}{1 + \log \left(2\pi \sigma_0^2\right)},
\end{equation}
As the RHS is decreasing and the LHS is increasing, this equation does not make sense. In addition, the RHS is not bounded between $(0,1)$, to which we conclude the heuristic construct by Austin et al.~is not a principled design for the OU process.

\printbibliography

@book{vanKampen,
  title = {Stochastic {{Processes}} in {{Physics}} and {{Chemistry}}},
  author = {van Kampen, N G},
  options = {useprefix=true},
  date = {2007},
  publisher = {{Elsevier Science B.V.}},
  location = {{Amsterdam}}
}

@book{gardinerStochasticMethodsHandbook2009,
  title = {Stochastic Methods: A Handbook for the Natural and Social Sciences},
  shorttitle = {Stochastic Methods},
  author = {Gardiner, Crispin W.},
  date = {2009},
  series = {Springer Series in Synergetics},
  edition = {4th ed},
  number = {13},
  publisher = {{Springer}},
  location = {{Berlin Heidelberg}},
  langid = {english},
  pagetotal = {447},
  note = {Includes bibliographical references and index. - Previous ed.: 2004}
}

@misc{nicholImprovedDenoisingDiffusion2021,
  title = {Improved {{Denoising Diffusion Probabilistic Models}}},
  author = {Nichol, Alex and Dhariwal, Prafulla},
  date = {2021-02-18},
  number = {arXiv:2102.09672},
  eprint = {2102.09672},
  eprinttype = {arxiv},
  primaryclass = {cs, stat},
  publisher = {{arXiv}},
  url = {http://arxiv.org/abs/2102.09672},
  urldate = {2022-09-27},
  archiveprefix = {arXiv},
  langid = {english}
}

@misc{songScoreBasedGenerativeModeling2021a,
  title = {Score-{{Based Generative Modeling}} through {{Stochastic Differential Equations}}},
  author = {Song, Yang and Sohl-Dickstein, Jascha and Kingma, Diederik P. and Kumar, Abhishek and Ermon, Stefano and Poole, Ben},
  date = {2021-02-10},
  number = {arXiv:2011.13456},
  eprint = {2011.13456},
  eprinttype = {arxiv},
  primaryclass = {cs, stat},
  publisher = {{arXiv}},
  url = {http://arxiv.org/abs/2011.13456},
  urldate = {2022-09-27},
  archiveprefix = {arXiv},
  langid = {english},
  note = {Comment: ICLR 2021 (Oral)}
}

@misc{hoDenoisingDiffusionProbabilistic2020,
  title = {Denoising {{Diffusion Probabilistic Models}}},
  author = {Ho, Jonathan and Jain, Ajay and Abbeel, Pieter},
  date = {2020-12-16},
  number = {arXiv:2006.11239},
  eprint = {2006.11239},
  eprinttype = {arxiv},
  primaryclass = {cs, stat},
  publisher = {{arXiv}},
  url = {http://arxiv.org/abs/2006.11239},
  urldate = {2022-09-27},
  archiveprefix = {arXiv},
  langid = {english}
}

@misc{sohl-dicksteinDeepUnsupervisedLearning2015,
  title = {Deep {{Unsupervised Learning}} Using {{Nonequilibrium Thermodynamics}}},
  author = {Sohl-Dickstein, Jascha and Weiss, Eric A. and Maheswaranathan, Niru and Ganguli, Surya},
  date = {2015-11-18},
  number = {arXiv:1503.03585},
  eprint = {1503.03585},
  eprinttype = {arxiv},
  primaryclass = {cond-mat, q-bio, stat},
  publisher = {{arXiv}},
  url = {http://arxiv.org/abs/1503.03585},
  urldate = {2022-09-27},
  archiveprefix = {arXiv},
  langid = {english}
}

@article{originalOU,
  title = {On the Theory of the Brownian Motion},
  author = {Uhlenbeck, G. E. and Ornstein, L. S.},
  journal = {Phys. Rev.},
  volume = {36},
  issue = {5},
  pages = {823--841},
  numpages = {0},
  year = {1930},
  month = {9},
  publisher = {American Physical Society},
  doi = {10.1103/PhysRev.36.823},
  url = {https://link.aps.org/doi/10.1103/PhysRev.36.823}
}

@inproceedings{campbell,
 author = {Campbell, Andrew and Benton, Joe and De Bortoli, Valentin and Rainforth, Thomas and Deligiannidis, George and Doucet, Arnaud},
 booktitle = {Advances in Neural Information Processing Systems},
 editor = {S. Koyejo and S. Mohamed and A. Agarwal and D. Belgrave and K. Cho and A. Oh},
 pages = {28266--28279},
 publisher = {Curran Associates, Inc.},
 title = {A Continuous Time Framework for Discrete Denoising Models},
 url = {https://proceedings.neurips.cc/paper_files/paper/2022/file/b5b528767aa35f5b1a60fe0aaeca0563-Paper-Conference.pdf},
 volume = {35},
 year = {2022}
}

@inproceedings{NEURIPS2020_92c3b916,
 author = {Song, Yang and Ermon, Stefano},
 booktitle = {Advances in Neural Information Processing Systems},
 editor = {H. Larochelle and M. Ranzato and R. Hadsell and M.F. Balcan and H. Lin},
 pages = {12438--12448},
 publisher = {Curran Associates, Inc.},
 title = {Improved Techniques for Training Score-Based Generative Models},
 url = {https://proceedings.neurips.cc/paper_files/paper/2020/file/92c3b916311a5517d9290576e3ea37ad-Paper.pdf},
 volume = {33},
 year = {2020}
}

@inproceedings{austin,
 author = {Austin, Jacob and Johnson, Daniel D. and Ho, Jonathan and Tarlow, Daniel and van den Berg, Rianne},
 booktitle = {Advances in Neural Information Processing Systems},
 editor = {M. Ranzato and A. Beygelzimer and Y. Dauphin and P.S. Liang and J. Wortman Vaughan},
 pages = {17981--17993},
 publisher = {Curran Associates, Inc.},
 title = {Structured Denoising Diffusion Models in Discrete State-Spaces},
 url = {https://proceedings.neurips.cc/paper_files/paper/2021/file/958c530554f78bcd8e97125b70e6973d-Paper.pdf},
 volume = {34},
 year = {2021}
}

@inproceedings{DDS,
  author       = {Francisco Vargas and
                  Will Sussman Grathwohl and
                  Arnaud Doucet},
  title        = {Denoising Diffusion Samplers},
  booktitle    = {The Eleventh International Conference on Learning Representations,
                  {ICLR} 2023, Kigali, Rwanda, May 1-5, 2023},
  publisher    = {OpenReview.net},
  year         = {2023},
  url          = {https://openreview.net/pdf?id=8pvnfTAbu1f},
}

%%%%%%%%%%%%%%%%%%%%%%%%%%%%%%%%%%%%%%%%%%%%%%%%%%%%%%%%%%%%

\end{document}